\documentclass[lettersize,journal]{IEEEtran}
\usepackage{amsmath,amsfonts}
\usepackage{algorithmic}
\usepackage{algorithm}
\usepackage{array}
\usepackage[caption=false,font=normalsize,labelfont=sf,textfont=sf]{subfig}
\usepackage{textcomp}
\usepackage{stfloats}
\usepackage{url}
\usepackage{verbatim}
\usepackage{graphicx}
\usepackage{cite}
\usepackage {makecell}
\usepackage{multirow}

\usepackage{wrapfig}  
\usepackage{gensymb}
\usepackage{booktabs}  
\usepackage{siunitx}  
\usepackage{arydshln} 
\usepackage[dvipsnames,table]{xcolor}

\usepackage{ulem}

\usepackage[pagebackref,breaklinks,colorlinks,citecolor=cvprblue]{hyperref}
\definecolor{cvprblue}{rgb}{0.21,0.49,0.74}
\usepackage{colortbl}

\begin{document}

\title{CDI3D: Cross-guided Dense-view Interpolation for\\ 3D Reconstruction
}

\author{
Zhiyuan Wu$^{1 *}$,
Xibin Song$^{2 \dagger}$,
Senbo Wang$^{2}$,
Weizhe Liu$^{2}$,
Jiayu Yang$^{2}$,
Ziang Cheng$^{2}$,
Shenzhou Chen$^{2}$,
Taizhang Shang$^{2}$,
Weixuan Sun$^{2}$,
Shan Luo$^{1}$, \textit{Senior Member, IEEE}, 
and Pan Ji$^{2}$

\thanks{$^1$Zhiyuan Wu and Shan Luo are with Department of Engineering, King's College London, Strand, London, WC2R 2LS, United Kingdom, \{zhiyuan.1.wu, shan.luo\}@kcl.ac.uk.}
\thanks{$^2$Xibin Song, Senbo Wang, Weizhe Liu, Jiayu Yang, Ziang Cheng, Shenzhou Chen, Taizhang Shang, Weixuan Sun, and Pan Ji are with Tencent XR Vision Labs, Shanghai, China. }
\thanks{*The contribution of Zhiyuan Wu was made during an internship at Tencent XR Vision Labs.}
\thanks{$^\dagger$Corresponding author}
}



\maketitle

\normalem

\begin{abstract}
3D object reconstruction from single-view image is a fundamental task in computer vision with wide-ranging applications. Recent advancements in Large Reconstruction Models (LRMs) have shown great promise in leveraging multi-view images generated by 2D diffusion models to extract 3D content. However, challenges remain as 2D diffusion models often struggle to produce dense images with strong multi-view consistency, and LRMs tend to amplify these inconsistencies during the 3D reconstruction process. Addressing these issues is critical for achieving high-quality and efficient 3D reconstruction. In this paper, we present CDI3D, a feed-forward framework designed for efficient, high-quality image-to-3D generation with view interpolation. To tackle the aforementioned challenges, we propose to integrate 2D diffusion-based view interpolation into the LRM pipeline to enhance the quality and consistency of the generated mesh. Specifically, our approach introduces a Dense View Interpolation (DVI) module, which synthesizes interpolated images between main views generated by the 2D diffusion model, effectively densifying the input views with better multi-view consistency. We also design a tilt camera pose trajectory to capture views with different elevations and perspectives. Subsequently, we employ a tri-plane-based mesh reconstruction strategy to extract robust tokens from these interpolated and original views, enabling the generation of high-quality 3D meshes with superior texture and geometry. Extensive experiments demonstrate that our method significantly outperforms previous state-of-the-art approaches across various benchmarks, producing 3D content with enhanced texture fidelity and geometric accuracy.
\end{abstract}

\begin{IEEEkeywords}
3D generation, multi-view diffusion, large reconstruction model
\end{IEEEkeywords}

\section{Introduction}
\label{sec:intro}
\IEEEPARstart{H}{umans} possess an exceptional capacity to envision the approximate 3D shape of an object from a single RGB image, drawing on their previous knowledge and experience to mentally reconstruct its geometry \cite{sun2023tcsvt1}.  Modeling this reconstruction process using computational methods has become a fascinating research topic in the field of computer vision, commonly referred to as image-to-3D generation. This technology has gained increasing importance in various applications, including virtual reality, gaming, and robotics, where the ability to infer 3D structures from 2D observations is critical to creating immersive environments and enabling machine interactions with the physical world \cite{pang2024envision3d}. One of the key reasons for humans' proficiency in image-to-3D generation lies in their ability to harness extensive prior knowledge about the structure, geometry, and appearance of objects. Inspired by this, researchers have sought to replicate this capability in machines by incorporating prior knowledge into computational models. With the rapid development of deep learning, significant progress has been made in this area. Deep learning models, with their ability to learn complex priors from large-scale datasets \cite{deitke2023objaverse, deitke2024objaversexl}, have become the dominant approach to tackle image-to-3D generation, aiming to mimic the human process of learning and reasoning to predict 3D shapes from images with increasing accuracy and realism. 

Recent mainstream deep learning-based 3D representations, such as voxels \cite{choy2016voxel1, fan2023tcsvt3}, point clouds \cite{fan2017pcd1, wang2021tcsvt2}, and implicit functions \cite{chen2019implicit1, yu2024tcsvt4}, have been widely studied and applied in various 3D reconstruction tasks. Although these paradigms have demonstrated distinct advantages across various applications, their fundamental limitations persist, such as high memory consumption for voxels, lack of connectivity in point clouds, and complex optimization requirements for implicit functions. In contrast, meshes have gained increasing attention due to their ability to explicitly encode both geometry and topology in a compact and structured form, making them particularly suitable for detailed surface reconstruction and widely used in practical applications. Previous works have explored the generation of meshes directly from images using end-to-end encoder-decoder architectures, where an encoder extracts image features and a decoder predicts the reconstructed mesh \cite{groueix2018mesh1, pan2019mesh3}. However, such methods face challenges, including high computational demands and difficulties in accurately modeling complex mesh geometry, often leading to suboptimal surface quality and limited generalization. 

\begin{figure*}[t!]
	\centering
	\includegraphics[width=\textwidth]{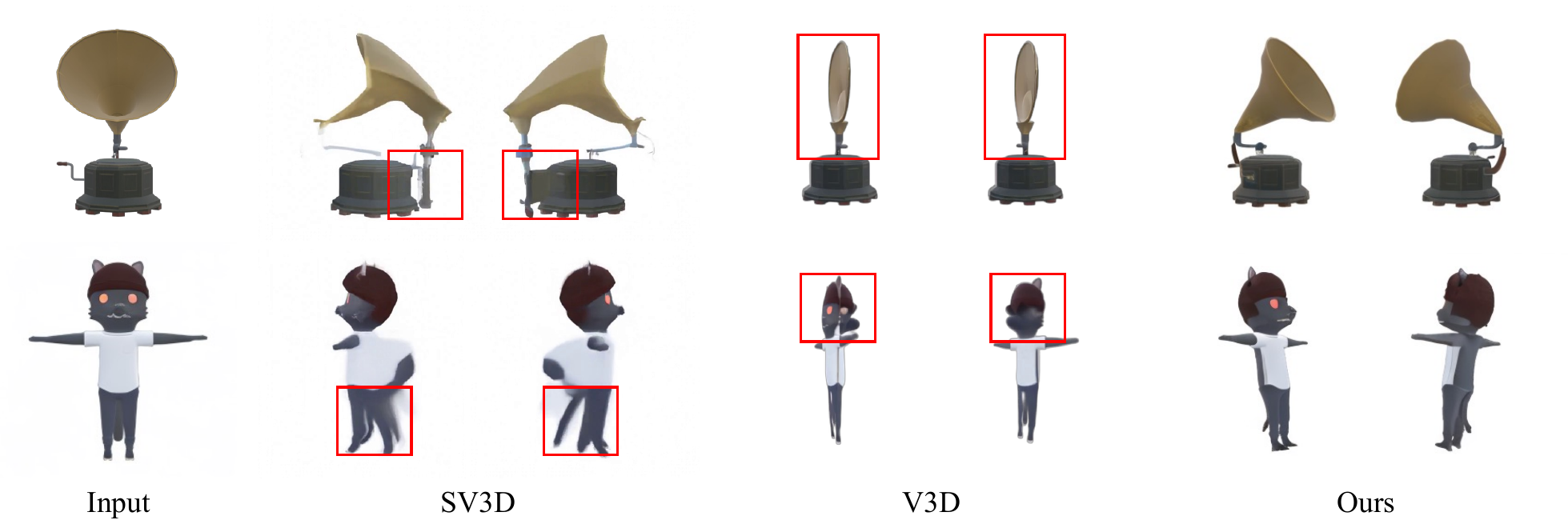}
    \caption{
    Qualitative comparisons between our DVI module and video diffusion methods in multi-view generation, including SV3D~\cite{voleti2024sv3d} and V3D~\cite{chen2024v3d}. Two generated images are shown here, and images generated by video diffusion networks show inconsistencies due to the lack of connectivity across frames. In contrast, our method ensures strong inter-frame connections, which significantly enhances the multi-view consistency of the generated images.
    }\label{fig.video}
\end{figure*}

Meanwhile, advancements in 2D diffusion models \cite{ho2020diffusion2d1, rombach2022stablediffusion} and Large Reconstruction Models (LRMs) \cite{hong2023lrm, li2023instant3d, tang2024lgm, wang2024crm, xu2024instantmesh} have opened new avenues for 3D content creation. Several works, including \cite{poole2023dreamfusion, lin2023magic3d, qian2023magic123, qiu2024richdreamer, chen2024it3d, chen2024text23d}, leverage 2D diffusion models to generate 3D content through a Score Distillation Sampling (SDS) pipeline, which optimizes a 3D representation by aligning its rendered 2D projections with the distribution learned by a pre-trained 2D diffusion model. An alternative paradigm involves generating multi-view images using 2D diffusion models, followed by applying reconstruction algorithms to infer 3D geometry and topology from these images \cite{liu2023zero123, shi2023mvdream, liu2023syncdreamer, wang2023imagedream, shi2023zero123++, long2024wonder3d}, resulting in more detailed geometry and texture reconstruction with improved performance. However, current state-of-the-art (SoTA) methods face notable limitations. Most approaches generate a limited number of multi-view images (typically four or six), which constrains the richness of geometric and textural details that can be captured. To address this, recent works such as \cite{blattmann2023svd, voleti2024sv3d, chen2024v3d} have proposed video diffusion strategies to directly expand the number of generated views. While these methods mark a step forward, they are often hindered by significant challenges, particularly multi-view inconsistency, where the generated views fail to align cohesively, leading to artifacts and degraded reconstruction quality, as illustrated in Fig.~\ref{fig.video} (SV3D~\cite{voleti2024sv3d} and V3D~\cite{chen2024v3d}). Furthermore, these approaches come with substantial computational overhead, requiring intensive GPU memory and extended training times, which severely limit their scalability and practical applicability in real-world scenarios. 

To address these limitations, we introduce \textbf{CDI3D}, \textit{i.e.}, \textbf{C}ross-guided \textbf{D}ense-view \textbf{I}nterpolation for
\textbf{3D} Reconstruction, a novel LRM-based image-to-3D framework that significantly improves 3D generation quality by leveraging dense 2D diffusion-based view interpolation. The core motivation behind CDI3D is to tackle the challenge of generating a large number of multi-view images by separating the process into two distinct steps, ensuring better constraints and consistency throughout the pipeline.
Specifically, CDI3D first employs a 2D diffusion model to generate $N$ main views (where $N$ is 4) from a single image, and then introduces an Dense View Interpolation (DVI) module to interpolate additional views with superior multi-view consistency, enriching geometric and textural details. By facilitating smooth and coherent transitions between neighboring main views, the DVI module effectively addresses the multi-view inconsistency issues that plague existing methods. Meanwhile, we design a tilt camera pose trajectory to capture views with different elevations and perspectives to maximize the coverage of the object’s surface. Finally, the interpolated views, along with the main views, are processed by a tri-plane-based mesh reconstruction model, which extracts robust tokens and reconstructs a high-quality 3D mesh with detailed geometry and texture. As shown in Fig. \ref{fig.video} (Ours), this design guarantees both multi-view consistency and high-quality image generation. 

Our main contributions are summarized as follows. \textbf{First}, we propose CDI3D, an LRM-based framework for high-quality 3D mesh reconstruction from a single image. By disentangling the process into two distinct stages: main-view generation and view interpolation, CDI3D leverages a 2D diffusion model to generate main views and introduces an Dense View Interpolation (DVI) module to synthesize additional views with superior multi-view consistency. This two-stage design ensures better constraints and smooth transitions between views, effectively addressing the multi-view inconsistency issues that hinder existing methods. \textbf{Second}, we develop the DVI module, which employs 2D diffusion-based multi-view interpolation to generate coherent intermediate views between neighboring main views. These interpolated views, combined with the main views, are processed by a tri-plane-based LRM to enhance the geometric and textural details of the reconstructed 3D mesh, which significantly improves the quality of both geometry and texture reconstruction, ensuring high-fidelity 3D generation. \textbf{Third}, we design a tilt camera pose trajectory that captures views from various angles and elevations, ensuring that the object is observed from a wider range of viewpoints and providing more complete input for the 3D mesh reconstruction process. Besides, we conduct extensive experiments on the Google Scanned Objects (GSO) dataset \cite{downs2022gso}, Objaverse dataset~\cite{deitke2023objaverse, deitke2024objaversexl} and diverse type of images collected from the web, demonstrating that CDI3D outperforms state-of-the-art methods both quantitatively and qualitatively in terms of 3D reconstruction accuracy, multi-view consistency, and visual fidelity.

\section{Related Works}
\subsection{3D Generation from Single Image}
3D generation from a single image is a long-standing and challenging problem in computer vision. Mainstream learning-based 3D representations include voxels \cite{choy2016voxel1, fan2023tcsvt3}, point clouds \cite{fan2017pcd1, wang2021tcsvt2}, and implicit functions \cite{chen2019implicit1, yu2024tcsvt4}. While these representations have shown promise, they often struggle to balance efficiency, accuracy, and scalability. Recently, meshes have emerged as a compelling alternative due to their structured and compact nature, which allows for explicit encoding of both geometry and topology. This makes them particularly advantageous for tasks requiring detailed and high-quality surface reconstruction.

Building on the advancements in 3D representations, the remarkable success of diffusion models in 2D image generation \cite{ho2020diffusion2d1, rombach2022stablediffusion} has inspired numerous works to explore their potential for 3D generation. By leveraging the powerful generative capabilities of diffusion models, researchers have begun to adapt these techniques to tackle the challenges of 3D content creation. A mainstream approach is directly training 3D generators using 3D ground truth \cite{zhou2021pvd, zheng2023las, wang2023rodin, gupta2023triplane1, shue2023triplane2}. For instance, \cite{zhou2021pvd} and \cite{zheng2023las} trained diffusion models to directly generate 3D voxels. In \cite{wang2023rodin} and \cite{shue2023triplane2}, a 3D-aware tri-plane diffusion model is introduced to produce NeRF \cite{mildenhall2021nerf} representations. Nonetheless, 3D diffusion methods tend to be time-consuming during optimization, and often show low quality in terms of texture and geometry. To deal with this, some studies have explored the utilization of 2D diffusion-based generators for 3D generation. DreamFusion \cite{poole2023dreamfusion} was the first to use 2D diffusion models to generate 3D content through a SDS pipeline. Building upon this work, \cite{lin2023magic3d, qian2023magic123, qiu2024richdreamer, chen2024it3d, chen2024text23d} have adopted the SDS pipeline to optimize various 3D representations such as NeRF, mesh, and gaussian splatting \cite{kerbl2023gaussian}. However, performing 3D generation tasks with 2D diffusion models often encounters issues related to multi-view inconsistency, indicating room for improvement. 

\subsection{Multi-view Diffusion Models}
Researchers have made great efforts to improve diffusion models in multi-view images generation. Zero123 \cite{liu2023zero123} was the first to encode camera pose as an additional condition to generate images from different specific views. On this basis, MVDream \cite{shi2023mvdream} replace self-attention in the Unet architecture with multi-view attention to facilitate multi-view consistency. Other works \cite{liu2023syncdreamer, wang2023imagedream, shi2023zero123++, long2024wonder3d} share a similar idea to generate 3D-aware and multi-view consistent 2D representations. These multi-view images can be further processed using techniques such as NeRF \cite{mildenhall2021nerf} and Gaussian Splatting \cite{kerbl2023gaussian} to obtain 3D representations. Nevertheless, existing multi-view diffusion models are constrained to generating a limited number of images from a single input image. Recent advancements \cite{blattmann2023svd, voleti2024sv3d, chen2024v3d} have sought to outcome this limitation by utilizing temporal priors in video diffusion models to boost the number of generated images. Despite these improvements, such strategies often neglect the connectivity between frames, resulting in inconsistencies and diminishing the quality of the generated 3D content. 

\subsection{Large Reconstruction Models}
The advent of large-scale 3D datasets \cite{deitke2023objaverse, deitke2024objaversexl} has significantly advanced the field of image-to-3D generation, bringing generalized reconstruction models to new heights. LRM \cite{hong2023lrm} was a pioneer that demonstrates the superiority of Transformer \cite{vaswani2017transformer} backbone in mapping image tokens to predict tri-plane NeRF under multi-view supervision. Building upon this foundation, Instant3D \cite{li2023instant3d} extends the input to multi-view images, largely enhancing the quality of image-to-3D generation through multi-view diffusion models. Inspired by Instant3D, subsequent methods such as LGM \cite{tang2024lgm} and GRM \cite{xu2024grm} further refine it by replacing NeRF representations with 3D Gassian Splatting \cite{kerbl2023gaussian} to improve the rendering efficiency. Recently, CRM \cite{wang2024crm} an InstantMesh \cite{xu2024instantmesh} takes advantage of FlexiCubes \cite{shen2023flexicubes} to improve both efficiency and quality of image-to-3D generation.

\begin{figure*}[t!]
	\centering
	\includegraphics[width=\textwidth]{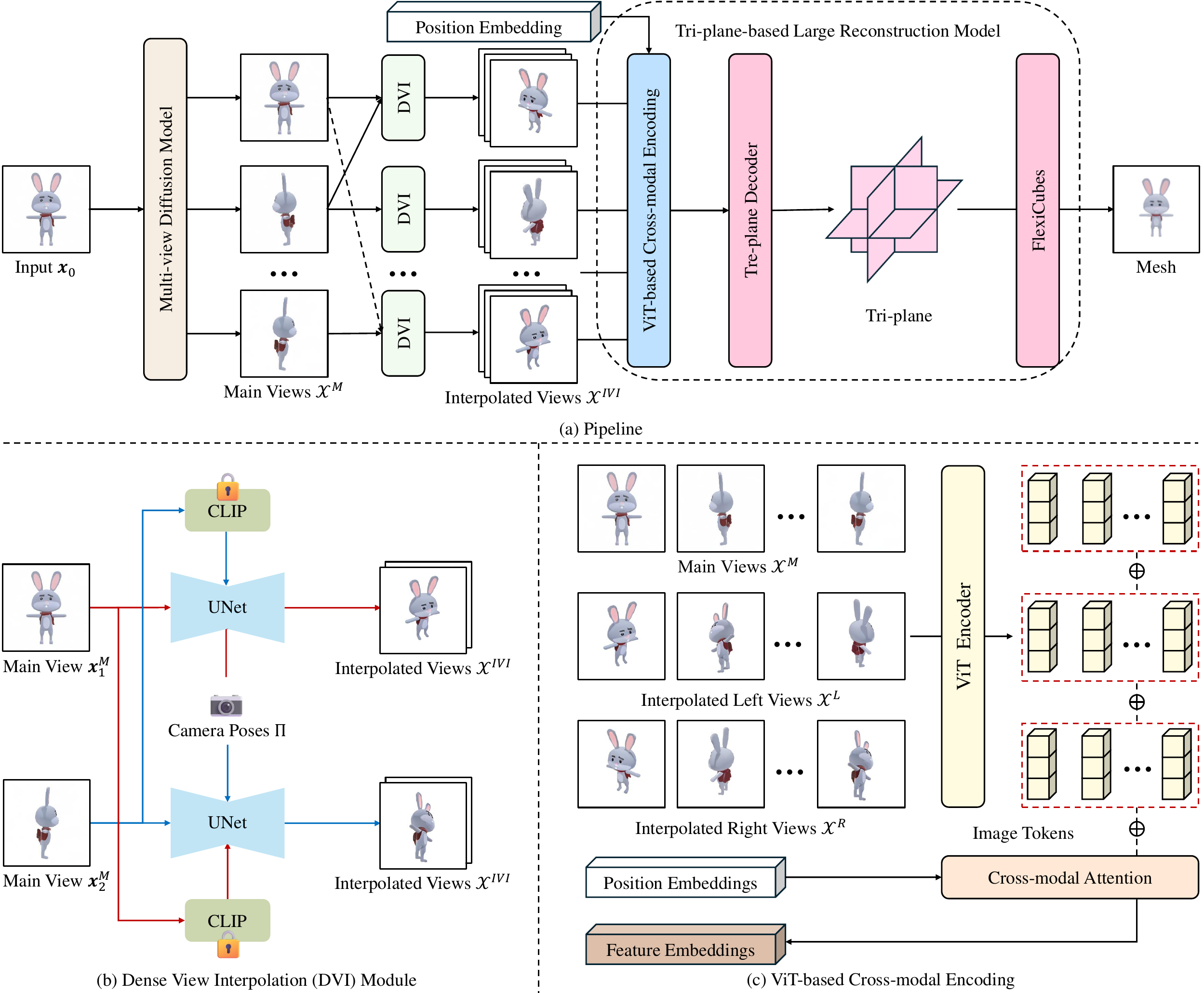}
    \caption{
    (a) The pipeline of our proposed CDI3D. Starting with a single image, CDI3D first generates main views using a multi-view diffusion model. (b) Interpolated views are then obtained from these main views using DVI module. (c) The images are processed through a ViT to extract feature embeddings, which are then used to generate a high-quality 3D mesh utilizing a tri-plane-based large reconstruction model. 
    }\label{fig.pipeline}
\end{figure*}

\section{Our Approach}
As illustrated in Fig.~\ref{fig.pipeline} (a), given a single input image $\boldsymbol{x}_0$, the architecture of our proposed CDI3D consists of 4 primary components: 1) a multi-view diffusion model to generate main multi-view images, 2) an \textbf{D}ense \textbf{V}iew \textbf{I}nterpolation (DVI) module to perform view interpolation between any two neighbouring views, 3) a tilt camera pose trajectory to capture views with different elevations and perspectives, and 4) a tri-plane based large reconstruction model to reconstruct a high-quality 3D mesh. The details of each component are elaborated below.

\subsection{Multi-view Diffusion Model}
In this paper, we follow \cite{long2024wonder3d} to train a four-view generation model based on multi-view 2D diffusion, which takes a single image as input, and generate outputs from four viewpoints (front, right, back, and left) to maximize multi-view consistency.

\subsection{Dense View Interpolation (DVI Module)}
Building upon main views generated by the multi-view diffusion model, we perform view interpolation through our DVI module. As depicted in Fig. \ref{fig.pipeline} (b), given two adjacent main view images $\boldsymbol{x}^M_{1}$ and $\boldsymbol{x}^M_{2} \in \mathbb{R}^{H \times W \times 3}$, our objective is to learn a model $f$ that synthesizes any interpolated image $\boldsymbol{x}_i$, along with their corresponding camera poses $\Pi = \{ \boldsymbol{\pi}^M_{1}, \boldsymbol{\pi}_i, \boldsymbol{\pi}^M_{2} \}$. Here $\boldsymbol{\pi} = [\boldsymbol{R}, \boldsymbol{T}]$, where $\boldsymbol{R} \in \mathbb{R}^{3 \times 3}$ and $\boldsymbol{T} \in \mathbb{R}^3$. This relationship can be formulated as follows: 
\begin{equation}
    \boldsymbol{x}_i = f(\boldsymbol{x}^M_{1}, \boldsymbol{x}^M_{2}, \Pi).
\end{equation}

Most multi-view diffusion architectures \cite{liu2023zero123, long2024wonder3d} employ the latent diffusion denoising strategy \cite{rombach2022stablediffusion}. In our view interpolation setting where two main views are input, one view is designated as the reference image $\boldsymbol{x}_i^{Ref}$, and the other as the condition image $\boldsymbol{x}_i^{Cond}$, so the adapted objective of the latent diffusion denoising process in our DVI module can be expressed as: 
\begin{equation}
    L_{DVI} := \mathbb{E}_{\boldsymbol{z} \sim \mathcal{E}(\boldsymbol{x}_i^{Ref}), t, \epsilon \sim \mathcal{N}(0,1)} \left\| \epsilon - \epsilon_{\theta}(\boldsymbol{z}_t, t, \mathcal{C}(\boldsymbol{x}_i^{Cond}, \boldsymbol{\pi}_i)) \right\|_2^2, 
\end{equation}
where $L_{DVI}$ represents the loss function in the DVI module that guides the latent diffusion process to generate interpolated views. In the formulation, $\mathcal{C}(\boldsymbol{x}_i^{Cond}, \boldsymbol{\pi}_i)$ represents the condition embedding of the condition view and the relative camera pose. The inference model $f$ is optimized to perform iterative denoising from $\boldsymbol{z}_T$ by training the model $\epsilon_{\theta}$ \cite{rombach2022stablediffusion}. Specifically, $\boldsymbol{z}_T$ is obtained by channel-concatenating $\boldsymbol{x}^{Ref}$. Following \cite{liu2023zero123}, a CLIP \cite{radford2021clip} embedding of $\boldsymbol{x}_i^{Cond}$ is concatenated with $\boldsymbol{\pi}_i$. This ensures that the generated interpolated images maintain multi-view consistency with both $\boldsymbol{x}^{Ref}$ and $\boldsymbol{x}^{Cond}$, which benefits stability of view interpolation. 

Given the varying camera poses of each interpolated view, some views are positioned closer to $\boldsymbol{x}^M_{1}$ while others are nearer to $\boldsymbol{x}^M_{2}$. To ensure a balanced distribution and multi-view consistency, for $\boldsymbol{x}_i$, the reference and condition views can be expressed as follows:
\begin{equation}
    {[\boldsymbol{x}^{Ref}_i, \boldsymbol{x}^{Cond}_i]} = \left\{
    \begin{aligned}
        {[\boldsymbol{x}^M_{1}, \boldsymbol{x}^M_{2}]}, & \quad \text{if } i \leq \frac{n}{2}, \\
        {[\boldsymbol{x}^M_{2}, \boldsymbol{x}^M_{1}]}, & \quad \text{if } i > \frac{n}{2},
    \end{aligned}
    \right.
\end{equation}
where n represents the number of interpolated images. Better constraints are provided in the view interpolation process, thus better results can be expected. In our implementation, we set $n$ to $2$, empirically. 

In DVI module, two main views are employed as reference and condition to improve the consistency and stability of the interpolated images. The consistent interpolated images effectively supplement missing views, thereby enriching the detail during model reconstruction. We provide more analysis in the experiment section.

\subsection{Tilt Camera Pose Trajectory}
\begin{figure}[t!]
  \centering  
  \includegraphics[width=0.45\textwidth]{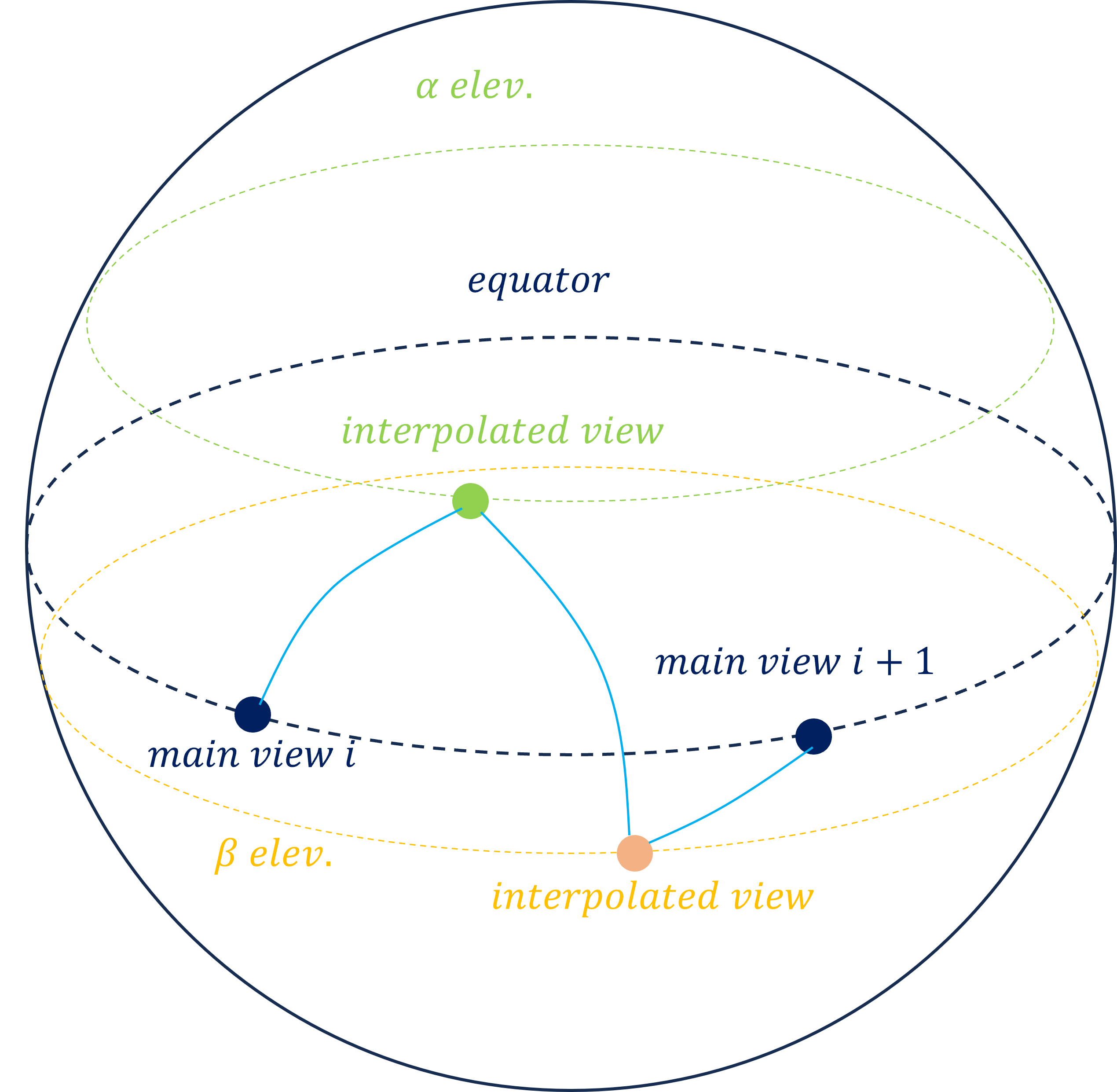}  
  \caption{
    Tilt camera pose trajectory design with elevations. 
  } \label{fig.elevation}
\end{figure}

To achieve a comprehensive 3D reconstruction of objects, it is crucial to design a camera trajectory that captures views from diverse angles and elevations. In CDI3D, we introduce a tilt camera pose trajectory with elevations that is carefully crafted to maximize the coverage of the object’s surface while ensuring consistency across interpolated views. As illustrated in Figure \ref{fig.elevation}, the trajectory is designed to include both horizontal and vertical variations, allowing the system to capture views with different elevations and perspectives. 

For any two adjacent main views $\boldsymbol{x}^M_{i}$ and $\boldsymbol{x}^M_{i+1}$, the DVI module generates interpolated views along a smooth trajectory defined by their camera poses. By incorporating elevation changes, the trajectory ensures that the object is observed from a wider range of viewpoints, effectively reducing occlusions and improving reconstruction quality. This trajectory design enriches the diversity of viewpoints, providing more complete input for the 3D mesh reconstruction process. Further analysis of its effectiveness is provided in the experiment section. 

\subsection{Tri-plane-based mesh reconstruction}
We train a robust tri-plane-based reconstruction model to obtain high-quality mesh from the multiple generated images. As illustrated in Fig. \ref{fig.pipeline} (c), for every two adjacent main images $\boldsymbol{x}^M_1$ and $\boldsymbol{x}^M_2$, we generate a sequence of interpolated images $\mathcal{X}^{DVI} = \{ \boldsymbol{x}_1, \dots, \boldsymbol{x}_n \}$ through our DVI module. Consequently, for each main view $\boldsymbol{x}^M_i$ in the set of sparse-view main images $\mathcal{X}^M = \{ \boldsymbol{x}^M_1, \dots, \boldsymbol{x}^M_N \}$ that generated by multi-view diffusion model, where $N$ represents the number of main views, we have interpolated images on its left and right: $\mathcal{X}^L = \{ \boldsymbol{x}^L_1, \dots, \boldsymbol{x}^L_n \}$ and $\mathcal{X}^R = \{ \boldsymbol{x}^R_1, \dots, \boldsymbol{x}^R_n \}$, respectively. Following general large reconstruction models \cite{hong2023lrm, li2023instant3d, xu2024instantmesh, wei2024meshlrm, xu2024grm}, we employ a Vision Transformer (ViT) $\mathcal{V}$ \cite{dosovitskiy2020vit} to extract image tokens from $\mathcal{X}^M$ and their corresponding $\mathcal{X}^L$ and $\mathcal{X}^R$ and add them to a position embedding through residual connection. This process can be written as follows:
\begin{equation} \label{eq.vit}
    \boldsymbol{f}^F = \boldsymbol{p} + \mathcal{A}_{cm} (\boldsymbol{p}, 
   \mathcal{V} (\mathcal{X}^M) \oplus \mathcal{V} (\mathcal{X}^L) \oplus \mathcal{V} (\mathcal{X}^R)), 
\end{equation}
where $\boldsymbol{f}^F$ represents the fused feature embeddings, $\boldsymbol{p}$ represents the initial position embedding, $\oplus$ represents channel-wise concatenation, and $\mathcal{A}_{cm}$ represents a cross-modal attention operation, defined as:
\begin{equation}
    \mathcal{A}_{cm}(\boldsymbol{p}, \boldsymbol{f}) = softmax(\frac{\boldsymbol{q}\boldsymbol{k}^T}{\sqrt{d}}) \cdot \boldsymbol{v}, 
\end{equation}
with 
\begin{equation}
    \boldsymbol{q} = \boldsymbol{w}_q \cdot \boldsymbol{p}, \quad \boldsymbol{k} = \boldsymbol{w}_k \cdot \boldsymbol{f}, \quad \boldsymbol{v} = \boldsymbol{w}_v \cdot \boldsymbol{f}, 
\end{equation}
where $\boldsymbol{w}$ denotes learnable projection matrices \cite{vaswani2017transformer, dosovitskiy2020vit}. In this learnable way, the main and interpolated image tokens are fused via residual connection to enhance multi-view consistency. Subsequently, following InstantMesh \cite{xu2024instantmesh}, we decode $\boldsymbol{f}^F$ to obtain a tri-plane representation, and reconstruct the final mesh through FlexiCubes \cite{shen2023flexicubes}. Thanks to our DVI module, more multi-view consistent image tokens are provided, bringing more details related to texture and geometry, thus resulting in a high-quality reconstructed mesh. 

The loss function for mesh reconstruction can be expressed as follows:
\begin{equation}  
\begin{split}  
    \mathcal{L} = &\mathcal{L}_{rgb} + \lambda_{depth}\mathcal{L}_{depth} + \lambda_{normal}\mathcal{L}_{normal} \\
    &+ \lambda_{mask}\mathcal{L}_{mask} + \lambda_{lpips}\mathcal{L}_{lpips} + \lambda_{reg}\mathcal{L}_{reg}, 
\end{split}  
\end{equation}
where $\mathcal{L}_{rgb}$, $\mathcal{L}_{depth}$, $\mathcal{L}_{normal}$, and $\mathcal{L}_{mask}$ refer to the loss of RGB images, depth, normal, and mask maps of the reconstructed mesh, and $\mathcal{L}_{lpips}$ and $\mathcal{L}_{reg}$ refer to LPIPS \cite{zhang2018lpips} and regression loss, respectively, with $\lambda_{lpips}=2.0$, $\lambda_{mask}=1.0$, $\lambda_{depth}=0.5$, $\lambda_{normal}=0.2$, $\lambda_{reg}=0.01$. 
Readers may refer to \cite{xu2024instantmesh} for more details. 

\section{Experiments}
In this section, we conduct a series of experiments quantitatively and qualitatively to evaluate the performance of our proposed CDI3D. We compare CDI3D against SoTA multi-view and image-to-3D baseline methods. Additionally, we perform ablation studies to validate the effectiveness and expand-ability of our proposed DVI module. 

\begin{figure*}[t!]
  \centering  
  \includegraphics[width=0.99\textwidth]{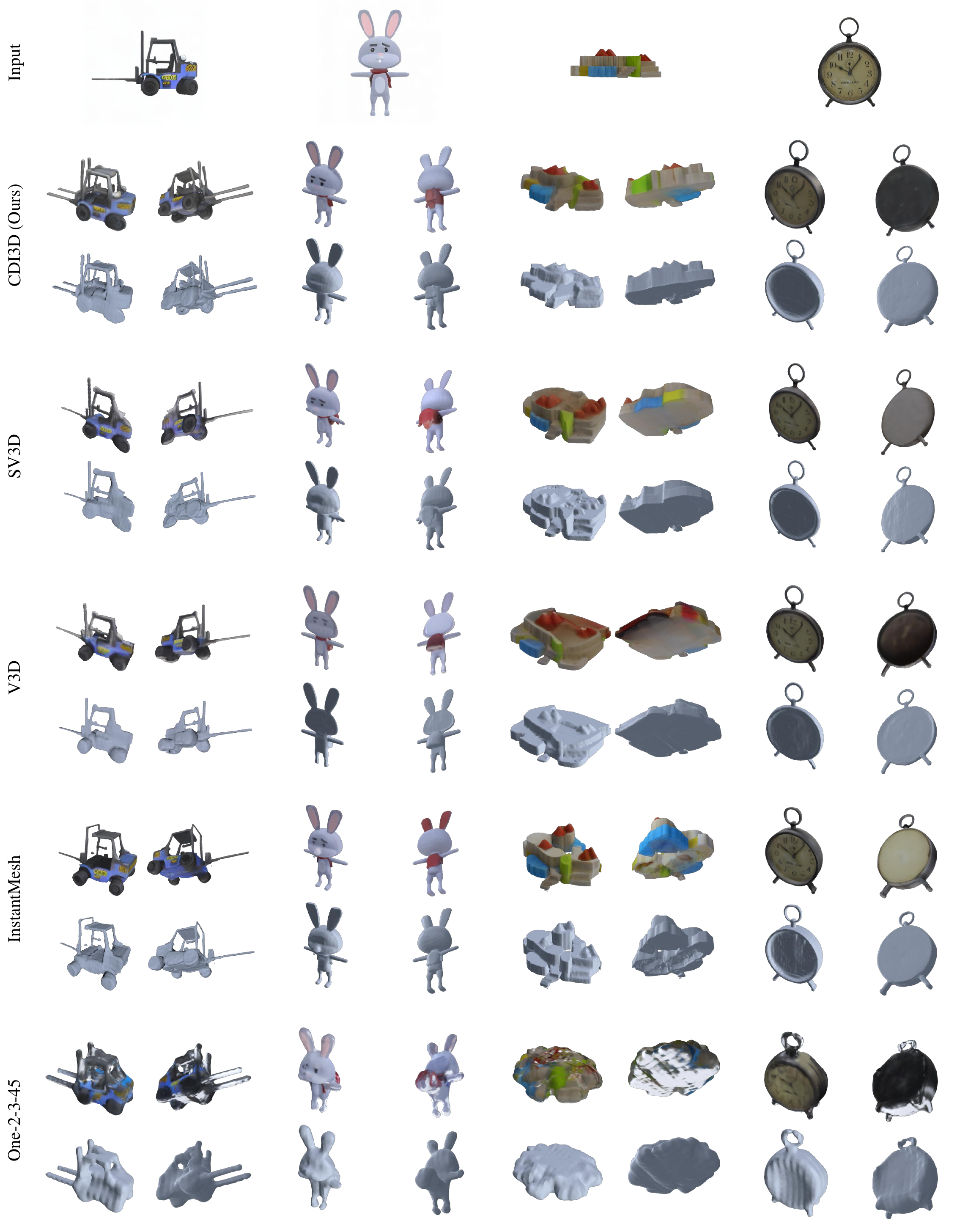}  
  \caption{Qualitative 3D mesh results generated by CDI3D demonstrate better geometry and texture compared to other baselines, where the forklift and rabbit are from Objaverse dataest, while the others are from GSO dataset.} \label{fig.mesh}
\end{figure*}

\subsection{Experimental Settings}

\textbf{Dataset. }Following prior research \cite{liu2023zero123, liu2023syncdreamer, long2024wonder3d}, we utilize the Google Scanned Objects dataset \cite{downs2022gso}, Objaverse dataset~\cite{deitke2023objaverse, deitke2024objaversexl} and diverse types of images from website for our evaluation, which encompasses a diverse array of common everyday objects. For the evaluation phase, we choose 30 representative objects ranging from everyday items to animals.

\textbf{Implementation Details. }
Our model is trained on the LVIS subset of the Objaverse dataset \cite{deitke2023objaverse}, consisting of approximately 30,000+ objects after a thorough cleanup process. For image interpolation, we fine-tune our DVI module starting from Wonder3D \cite{long2024wonder3d}, which has previously been fine-tuned for multi-view generation. During the fine-tuning process, we resize the image to 256 $\times$ 256 and employ a batch size of 128. This fine-tuning is performed for 10,000 steps. For mesh reconstruction, starting from InstantMesh \cite{xu2024instantmesh}, we fine-tune the model for 30,000 steps with a total batch size of 4. We use eight Nvidia A100 40GB in this paper. In both fine-tuning processes, we remain the original optimizer settings and $\epsilon$-prediction strategy. 

\textbf{Baselines and Metrics. }
For comparative analysis, we adopt One-2-3-45 \cite{liu2024one2345}, SyncDreamer \cite{liu2023syncdreamer}, Wonder3D \cite{long2024wonder3d}, Magic123 \cite{qian2023magic123}, LGM \cite{tang2024lgm}, InstantMesh \cite{xu2024instantmesh}, V3D \cite{chen2024v3d}, and SV3D \cite{voleti2024sv3d} as our baselines to evaluate the quality of the generated mesh. We also adopt V3D and SV3D as our baselines to evaluate the quality of novel view synthesis of our DVI module in orbiting view generation. 

To evaluate the geometry quality for 3D textured mesh generation, Chamfer Distances, Volume IoU, and F-score metrics are utilized.
To evaluate novel view synthesis (NVS) and the texture quality for 3D texutred mesh generation, we employ the PSNR, SSIM \cite{wang2004ssim}, and LPIPS \cite{zhang2018lpips} as the evaluation metrics. We also evaluate the GPU memory usage in orbiting view generation. 

\subsection{3D Textured Mesh Generation}
The quantitative results are summarized in Tabs. \ref{tab.3d} and \ref{tab.3d_nvs}, where our CDI3D outperforms all baseline methods in terms of both geometric and texture quality metrics. For mesh texture evaluation, we render $24$ images at $512 \times 512$ resolution, capturing meshes at elevation angles of $0^\circ$, $15^\circ$, and $30^\circ$, with 8 images evenly distributed around a full $360^\circ$ rotation for both generated and ground-truth meshes. Among the baseline models, though InstantMesh demonstrates better performance in geometry quality, and SV3D demonstrates better performance in texture quality, our results outperform these SOTAs in both geometry and texture. Based on high-quality main view results, the diverse detail acquisition from the DVI module enables the reconstruction model to capture comprehensive geometric and texture information, which is proved in ablation studies in Sec. \ref{exp.ablation}. 

Qualitative comparisons in Fig. \ref{fig.mesh} including images collected from Objaverse dataset~\cite{deitke2023objaverse, deitke2024objaversexl} and the GSO dataset~\cite{downs2022gso}. Our consistent view interpolation approach enriches image tokens within the reconstruction model, providing more features with good multi-view consistency, therefore, comparing with SOTAs, more smooth geometry and visual appealing textures can be obtained by our approach.
We also present rendered \ang{360} reconstruction dense images to better show the details of our mesh results, as shown in Fig. \ref{fig.elev_new}.

\begin{table}[t!]  
\centering  
\footnotesize
\caption{  
    Quantitative comparison for geometry quality between our method and baselines for 3D textured mesh generation. We report Chamfer Distance, Volume IoU and F-score on the GSO dataset. The best results are shown in bold font.
} \label{tab.3d}
\begin{tabular}{l c c c} 
    \toprule  
    Method & Chamfer Dist. $\downarrow$ & Vol. IoU $\uparrow$ & F-Sco. $\uparrow$ \\
    \midrule
    One-2-3-45 & 0.0172 & 0.4463 & 0.7219 \\
    SyncDreamer & 0.0140 & 0.3900 & 0.7574 \\
    Wonder3D & 0.0186 & 0.4398 & 0.7675 \\
    Magic123 & 0.0188 & 0.3714 & 0.6066 \\
    LGM & 0.0117 & 0.4685 & 0.6869 \\
    InstantMesh & 0.0103 & 0.5712 & 0.7121 \\
    V3D & 0.0143 & 0.4660 & 0.6234 \\
    SV3D & 0.0142 & 0.4949 & 0.6529 \\
    \textbf{Ours} & \textbf{0.0101} & \textbf{0.6399} & \textbf{0.7765} \\
    \bottomrule
\end{tabular}
\end{table}

\begin{table}[t!]  
\centering 
\footnotesize
\caption{  
    Quantitative comparison for texture quality between our method and baselines for 3D textured mesh generation. We report PSNR, SSIM \cite{wang2004ssim}, LPIPS \cite{zhang2018lpips} on the GSO dataset. The best results are shown in bold font.
} \label{tab.3d_nvs}
\begin{tabular}{l c c c} 
    \toprule  
    Method & PSNR $\uparrow$ & SSIM $\uparrow$ & LPIPS $\downarrow$ \\
    \midrule
    One-2-3-45 & 13.93 & 0.8084 & 0.2625 \\
    SyncDreamer & 14.00 & 0.8165 & 0.2591 \\
    Wonder3D & 13.31 & 0.8121 & 0.2554 \\
    Magic123 & 12.69 & 0.7984 & 0.2442 \\
    LGM & 13.28 & 0.7946 & 0.2560 \\
    InstantMesh & 17.66 & 0.8053 & 0.1517 \\
    V3D & 17.60 & 0.8115 & 0.1520 \\
    SV3D & 17.76 & 0.8173 & 0.1517 \\
    \textbf{Ours} & \textbf{18.32} & \textbf{0.8230} & \textbf{0.1397} \\
    \bottomrule
\end{tabular}
\end{table}

\subsection{Novel View Synthesis}
We benchmark the novel view synthesis capabilities of our DVI module against video diffusion-based baselines in orbiting view generation, where 12 views are selected along a horizontal orbiting trajectory. Quantitative results are presented in Tab. \ref{tab.mv_nvs}. Our approach effectively employ two main views as reference and condition, thus improving the consistency and stability of the interpolated images. As shown in Tab.~\ref{tab.mv_nvs}, it is also worth mentioning that our DVI module requires a much lower memory cost for inference compared to video diffusion-based methods, as we generate views by two steps. 


\begin{table*}[t!]  
\centering  
\footnotesize
\caption{  
    Quantitative comparison between our method and video diffusion-based methods for novel view synthesis in orbiting view generation. We select 12 views along a horizontal orbiting trajectory and report PSNR, SSIM \cite{wang2004ssim}, LPIPS \cite{zhang2018lpips}, GPU memory usage on the GSO dataset. The best results are shown in bold font.
} \label{tab.mv_nvs}
\begin{tabular}{l c c c c c} 
    \toprule  
    Method & PSNR $\uparrow$ & SSIM $\uparrow$ & LPIPS $\downarrow$ & Memory(MiB) & Inference Time (s) $\downarrow$\\
    \midrule  
    V3D & 16.37 & 0.796 & 0.173 & 39786 & 85.198 \\
    SV3D & 17.12 & 0.801 & 0.185 & 39014 & 31.893 \\
    \textbf{Ours} & \textbf{17.38} & \textbf{0.803} & \textbf{0.159} & \textbf{9686} & \textbf{14.324} \\
    \bottomrule  
\end{tabular}
\end{table*}

\begin{table*}[t!]  
\centering  
\footnotesize
\caption{  
    Quantitative results for texture and geometry quality of our method with different elevation angles for 3D textured mesh generation. We report Chamfer Distance, Volume IoU, F-score, PSNR, SSIM \cite{wang2004ssim}, LPIPS \cite{zhang2018lpips} on the GSO dataset. The best results are shown in bold font.
} \label{tab.trajectory}
\begin{tabular}{l c c c c c c} 
    \toprule  
    Method & Chamfer Dist. $\downarrow$ & Vol. IoU $\uparrow$ & F-Sco. $\uparrow$ & PSNR $\uparrow$ & SSIM $\uparrow$ & LPIPS $\downarrow$ \\
    \midrule
    baseline w/o DVI                      & 0.0186 & 0.4398 & 0.7675 & 13.31 & 0.8121 & 0.2554 \\
    w/o elev.                        & 0.0102 & 0.6299 & 0.7686 & 18.19 & 0.8222 & 0.1417 \\
    w/ +\ang{15} and -\ang{15} elev. & 0.0101 & 0.6380 & 0.7753 & \textbf{18.32} & \textbf{0.8230} & 0.1399 \\
    w/ +\ang{30} and -\ang{15} elev. & \textbf{0.0101} & 0.6353 & 0.7734 & 18.27 & 0.8229 & \textbf{0.1397} \\
    w/ +\ang{30} and -\ang{30} elev. & 0.0101 & \textbf{0.6399} & \textbf{0.7765} & 18.28 & 0.8229 & 0.1405 \\
    \bottomrule
\end{tabular}
\end{table*}

\subsection{Inference time}

\textbf{Mesh reconstruction}: Our 3D mesh reconstructiont LRM part takes an average time of 1.464 seconds for inference, which is similar with InstantMesh that constructs meshes in an average time of 1.270 seconds. As shown in Fig.~\ref{fig.pipeline} (c) and Eq.~\ref{eq.vit}, all image tokens are concatenated for subsequent operations. We have a position embedding $p \in \mathbb{R}^{V, P, D}$ and a concatenated tensor $\mathcal{X} \in \mathbb{R}^{V, P, D}$, where $V$ represents the view number. $p$ serves as the query and $\mathcal{X}$ acts as the key in the cross-modal attention operation. 

It is noteworthy that our approach does not result in a computational time proportional to $V^2$. This is because we only increase the computational load in the image encoder's transformer (cross-modal attention) part. After this step, we employ a tri-plane transformer that concatenates and flattens features from all views, then decodes them into a fixed-shape tri-plane. Subsequent operations are based on this fixed-shape tri-plane, which does not increase computational overhead. Therefore, the additional computational time is primarily confined to the image encoder section, and the overall computational complexity is not proportional to $V^2$.

Besides, for the concatenated tensor $X \in \mathbb{R}^{V,P,D}$, though the theoretical time complexity of cross attention is $O((VP)^2, D)$, we use Pytorch and \cite{dao2023flashattention} is utilized to accelerate the attention process. 
Thus increase of $V$ bring acceptable time consuming, from $1.270$ seconds to $1.464$ seconds.

\textbf{Dense View Interpolation}: Tab.~\ref{tab.mv_nvs} demonstrates the inference time comparisons between our approach and video generation methods. Our DVI module takes 3.5s for a single view interpolation process. In our experiment, four interpolations are required, the total video generation time is approximately 14s.

\begin{figure*}[t!]
  \centering  
  \includegraphics[width=0.99\textwidth]{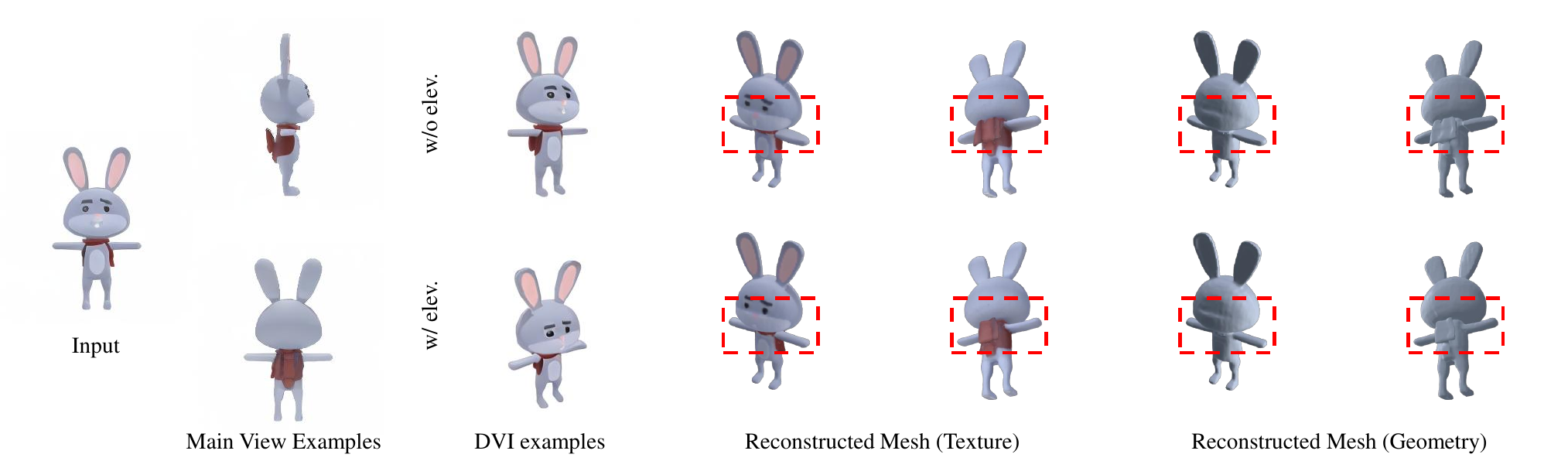}  
  \caption{
  DVI results of elevated camera trajectories and their corresponding reconstructed meshes. To highlight the differences, we present the results with and without a \ang{30} elevation. } \label{fig.elev}
\end{figure*}

\begin{figure*}[t!]
  \centering  
  \includegraphics[width=0.99\textwidth]{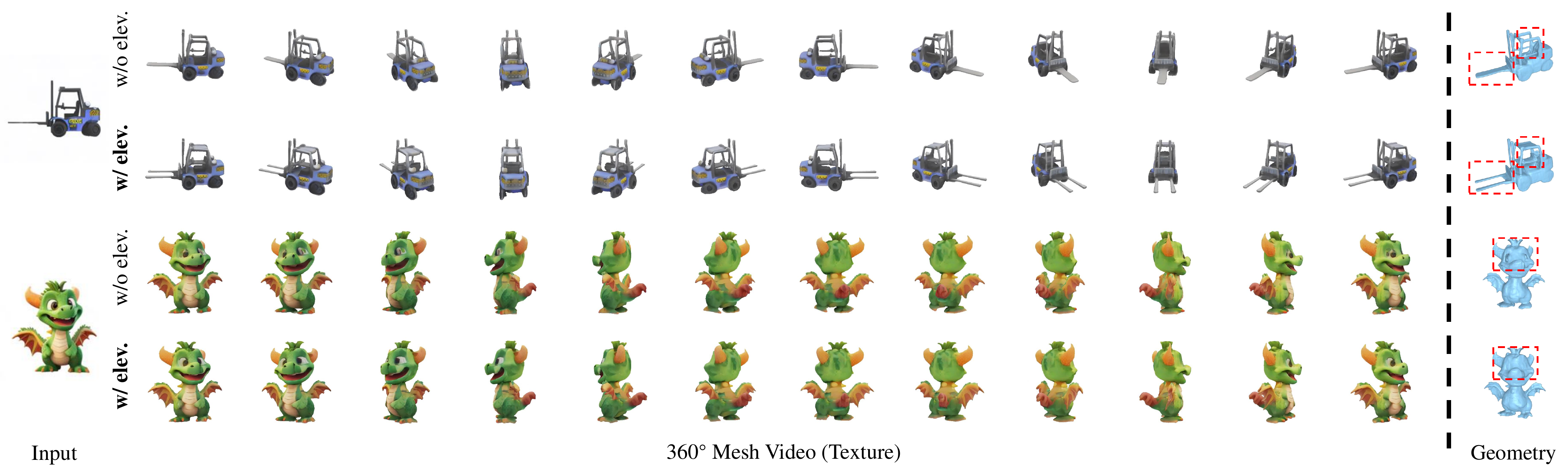}  
  \caption{Qualitative results on camera trajectories and \ang{360} reconstruction dense images.} \label{fig.elev_new}
\end{figure*}

\subsection{The Effectiveness of Tilt Camera Pose Trajectory}

We validate the effectiveness of our tilt camera pose trajectory design through both qualitative and quantitative evaluations. Tab.~\ref{tab.trajectory} illustrates the impact of varying elevation angles on camera pose trajectories, with representative examples shown in Fig.~\ref{fig.elev}. It can be observed that incorporating elevated camera trajectories (ranging from ±\ang{15} to ±\ang{30}) leads to noticeable improvements in both geometry and texture quality. This improvement is attributed to the richer detail diversity provided by elevated camera angles, as evidenced in the third column of Fig.~\ref{fig.elev}. Elevated trajectories allow the system to capture previously occluded or rarely observed areas, resulting in more complete and detailed reconstructions.

To further substantiate these findings, we provide quantitative results for different elevation angles in Tab.~\ref{tab.trajectory}. For comparison, we also include baseline results (Wonder3D w/o DVI) in the same table. The baseline results are obtained using Wonder3D without the DVI module. As shown, our designed camera trajectories, both with and without elevation, significantly outperform the baseline across all evaluation metrics, including Chamfer Distance, Volume IoU, PSNR, and LPIPS. Notably, trajectories with elevation achieve further improvements on all metrics. This is because trajectories without elevation already cover most of the object’s surface, while elevated trajectories capture additional details in rarely observed regions, leading to a more comprehensive reconstruction.

We also present more distinctive qualitative results in Fig.~\ref{fig.elev_new}, where differences in the reconstructed mesh geometry are highlighted in red. Elevated camera trajectories enable the DVI module to produce higher-quality meshes with finer details. For instance, the fork of the forklift and the eyes of the dragon are more complete and refined when using elevated trajectories. These results demonstrate that our designed camera trajectories, especially those with elevation, contribute positively to dense image generation and significantly enhance the quality of the final 3D mesh reconstruction.

\begin{table*}[t!]  
\centering  
\footnotesize
\caption{  
    Quantitative results for texture and geometry quality of our method with different number of interpolated number $n$ for 3D textured mesh generation. We report Chamfer Distance, Volume IoU, F-score, PSNR, SSIM \cite{wang2004ssim}, LPIPS \cite{zhang2018lpips} on the GSO dataset. The best results are shown in bold font.
} \label{tab.n_views}
\begin{tabular}{c c c c c c c} 
    \toprule  
    $n$ & Chamfer Dist. $\downarrow$ & Vol. IoU $\uparrow$ & F-Sco. $\uparrow$ & PSNR $\uparrow$ & SSIM $\uparrow$ & LPIPS $\downarrow$ \\
    \midrule
    baseline w/o DVI                      & 0.0186 & 0.4398 & 0.7675 & 13.31 & 0.8121 & 0.2554 \\
    $1$ & 0.0101 & 0.6297 & 0.7683 & 18.16 & 0.8209 & 0.1430 \\ 
    $2$ & 0.0102 & \textbf{0.6380} & \textbf{0.7753} & 18.19 & \textbf{0.8222} 
    & \textbf{0.1417} \\
    $3$ & \textbf{0.0101} & 0.6340 & 0.7719 & \textbf{18.21} & 0.8221 & 0.1424 \\
    \bottomrule
\end{tabular}
\end{table*}

\begin{figure*}[t!]
  \centering  
  \includegraphics[width=0.99\textwidth]{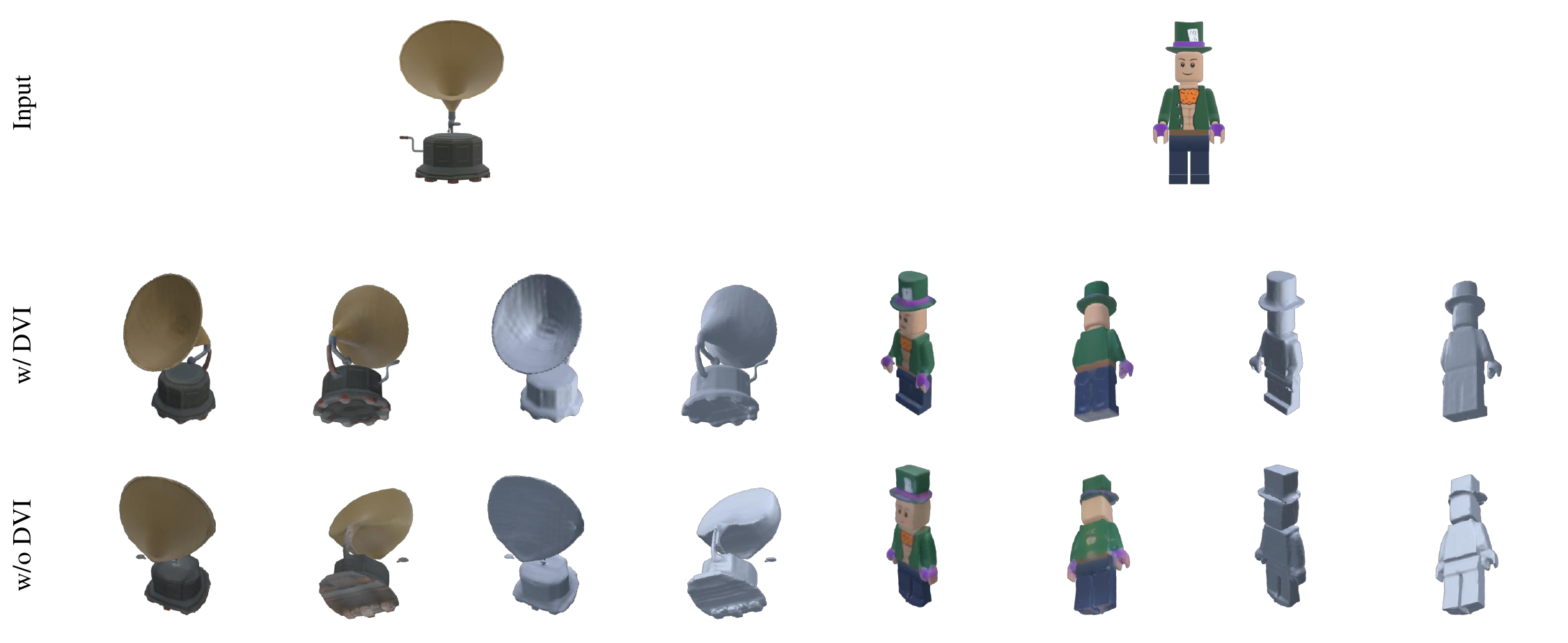}  
  \caption{We validate the effectiveness of our DVI module. It can be observed that view interpolation demonstrate better geometry and texture with more details. } \label{fig.aug}
\end{figure*}

\begin{figure*}[t!]
  \centering  
  \includegraphics[width=0.99\textwidth]{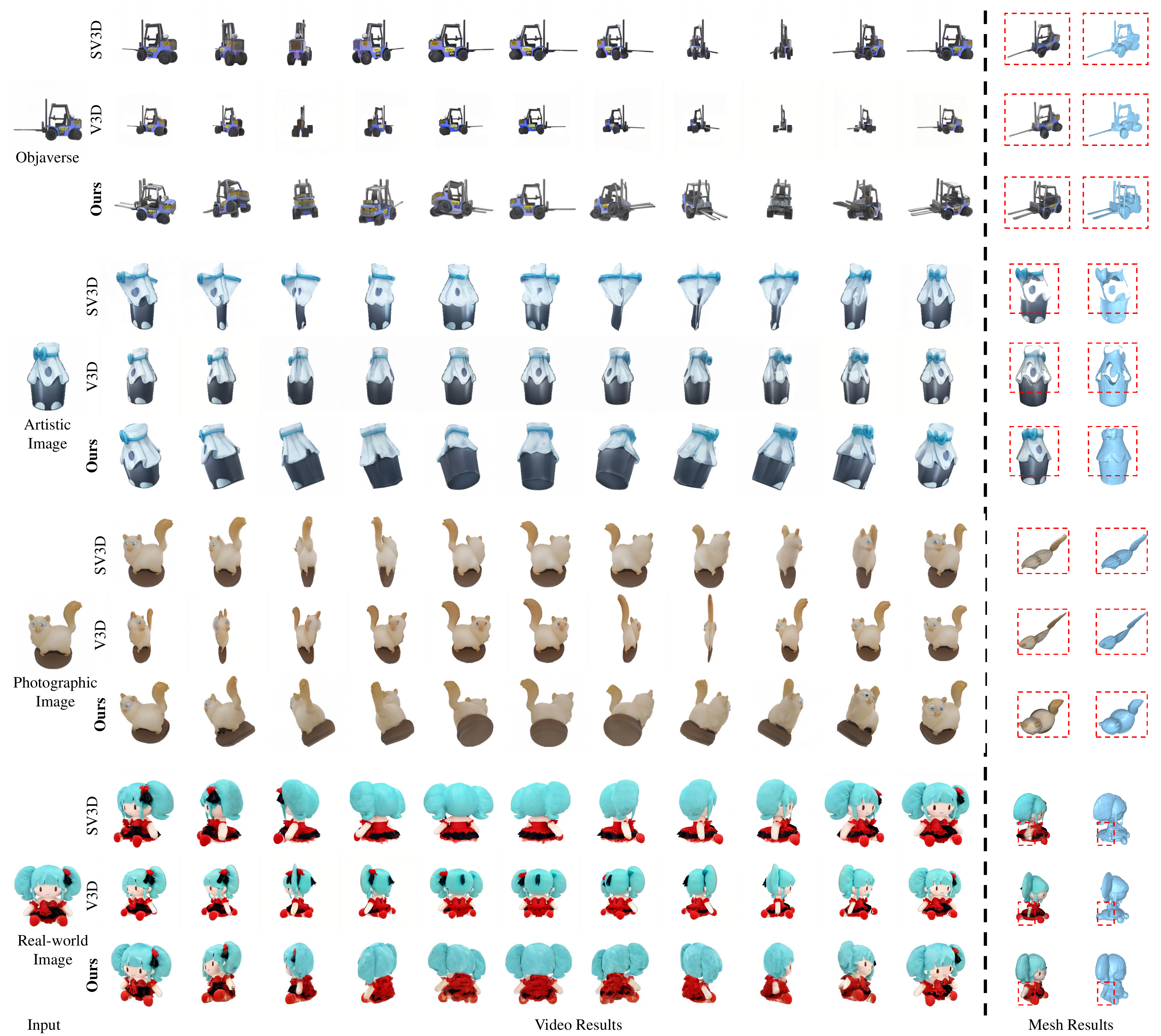}  
  \caption{Video and mesh results on out-of-distribution (OOD) data.} \label{fig.more_video}
\end{figure*}

\subsection{Ablation Study} \label{exp.ablation}
In this subsection, we conduct ablation study to validate the superiority of our architecture. 

\textbf{Dense View interpolation for LRM:} 
To evaluate the effectiveness of dense view interpolation in our LRM framework, we conduct ablation study with four views (front, right, back, and left) as input and tri-plane-based LRM for reconstruction. As illustrated in Fig. \ref{fig.aug}, with the DVI module generating interpolated images with superior multi-view consistency, our CDI3D reconstructs high quality meshes with more details and less breakage regarding geometry and texture, especially for objects with complicated geometry and texture. Meanwhile, as shown in Tab.~\ref{tab.trajectory}, the baseline results are obtained with Wonder3D since we use it as baseline without using DVI module. As shown in Tab.~\ref{tab.trajectory}, results with our DVI module with and without elevation all outperform baseline with large margins, which proves that all our designed camera trajectories work positively for dense image generation.

\textbf{Number of interpolation views:} We performed ablation studies to determine the optimal number $n$ of interpolated views. As illustrated in Table \ref{tab.n_views}, with setting $n=2$ yields the better performance in terms of both geometry and texture quality. Notably, when $n$ is set to $3$, similar results can be obtained comparing with $n$ = 2. Therefore, we set $n$ = 2 in our experiment. 

\subsection{Video and Mesh Results on OOD data}
We provide more out-of-distribution (OOD) visual results in Fig. \ref{fig.more_video} with different images as input, including both video and mesh results. We choose images from real-world, Objaverse dataset, and web (both artistic and photographic style), and our model is only trained with Objaverse dataset, which proves the generalization ability of our approach. 

As shown in the video results in Fig. \ref{fig.more_video}, better multi-view consistency images can be obtained by our approach, compared with other video-based methods, and differences in the mesh results are highlighted in red areas. For example, our method outperforms other video-based methods with more accurate geometry details in the forklift and cat, while SV3D and V3D show flattened results, treating three-dimensional objects as nearly two-dimensional objects. In the milk case, our approach effectively converts 2D artistic images into consistent multi-view images and intact meshes, maintaining shape consistency that others fail to achieve. Additionally, our method reconstructs more consistent details in the doll's arm, as highlighted in red areas, while other video-based methods result in texture blurring issue. 

\section{Limitation and Conclusion}
In this paper, we introduce CDI3D, a novel LRM-based image-to-3D framework to produce high-quality 3D content. Particularly, we propose an innovative multi-view diffusion-based DVI module to perform view interpolation, followed by a tri-plane-based mesh reconstruction to obtain the final mesh. Particularly, we design a tilt camera pose trajectory to capture views with different elevations and perspectives. Our experimental results indicate the superior performance of CDI3D, demonstrating its ability to generate 3D meshes with exceptional texture and geometric fidelity, compared to existing SoTA methods.

Based on our view interpolation strategy, we can achieve further view expansion of diverse trajectories by further applying the DVI module between the generated images. However, the performance of DVI module depends on the generation qualities of main view images in the first step. We believe improvements can be made by incorporating view super-resolution concept into multi-view diffusion at the feature level, which will be a primary focus of our future work. 

{
    \small
    \bibliographystyle{IEEEtran}
    \bibliography{main}
}

\vfill

\end{document}